\documentclass[sigconf,authorversion,noacm]{acmart}

\renewcommand\footnotetextcopyrightpermission[1]{}
\setcopyright{none}
\copyrightyear{}
\acmYear{}
\acmDOI{}
\acmConference[Document Intelligence Workshop at KDD]{The Second Document Intelligence Workshop at KDD}{2021}{Virtual Event}
\acmPrice{}
\acmISBN{}

\begin{document}
\fancyhead{}

\title{Detection Masking for Improved OCR on Noisy Documents}

\author{Daniel Rotman}
\authornote{Both authors contributed equally to this research}
\email{danieln@il.ibm.com}
\affiliation{%
  \institution{IBM Research - Haifa}
  \country{Israel}
}

\author{Ophir Azulai}
\authornotemark[1]
\email{ophir@il.ibm.com}
\affiliation{%
  \institution{IBM Research - Haifa}
  \country{Israel}
}

\author{Inbar Shapira}
\email{inbar_shapira@il.ibm.com}
\affiliation{%
  \institution{IBM Research - Haifa}
  \country{Israel}
}

\author{Yevgeny Burshtein}
\email{bursh@il.ibm.com}
\affiliation{%
  \institution{IBM Research - Haifa}
  \country{Israel}
}

\author{Udi Barzelay}
\email{udib@il.ibm.com}
\affiliation{%
  \institution{IBM Research - Haifa}
  \country{Israel}
}

\begin{abstract}
Optical Character Recognition (OCR), the task of extracting textual information from scanned documents is a vital and broadly used technology for digitizing and indexing physical documents.
Existing technologies perform well for clean documents, but when the document is visually degraded, or when there are non-textual elements, OCR quality can be greatly impacted, specifically due to erroneous detections.
In this paper we present an improved detection network with a masking system to improve the quality of OCR performed on documents.
By filtering non-textual elements from the image we can utilize document-level OCR to incorporate contextual information to improve OCR results.
We perform a unified evaluation on a publicly available dataset demonstrating the usefulness and broad applicability of our method.
Additionally, we present and make publicly available our synthetic dataset with a unique hard-negative component specifically tuned to improve detection results, and evaluate the benefits that can be gained from its usage.
\end{abstract}


\maketitle

\section{Introduction}

Detecting and recognizing words and characters in images is a cornerstone technology for information extraction in the visual domain \cite{ocr1}.
The difficulty of the task can often be divided into two categories:
Optical Character Recognition (OCR) is often incorporated when the image is a digitized (scanned) document consisting mostly of aligned text in standard fonts displayed on uniform backgrounds \cite{ocr2}.
For text appearing in natural images, Natural Scene Text recognition (NST) is often used, which incorporates advanced methods to overcome non-uniform backgrounds, non-standard fonts, and words appearing at odd angles or which have undergone spatial transformations \cite{nst}.
Often the benefits of using NST are less pronounced when dealing with scanned documents, and the extra computation power, especially when dealing with the large amounts of words appearing in a standard document, makes it less of a viable option.
Therefore in this work we focus mostly on OCR solutions for extracting text from documents.

One of the most broadly used solutions for OCR today is Tesseract \cite{tesseract}.
Tesseract is considered a commodity and the go-to solution when the given task is text extraction from documents.
The lightweight framework, multi-language support, ease of use, and open-source code provide an extremely useful resource.
However, Tesseract exhibits degraded results on documents exhibiting non-ideal conditions \cite{noise}.
Specifically, Tesseract tends toward false detections when there are noise artifacts or non-textual elements in the document such as logos, figures, and graphical elements.

\begin{figure*}[t]
  \centering
  \includegraphics[width=\linewidth]{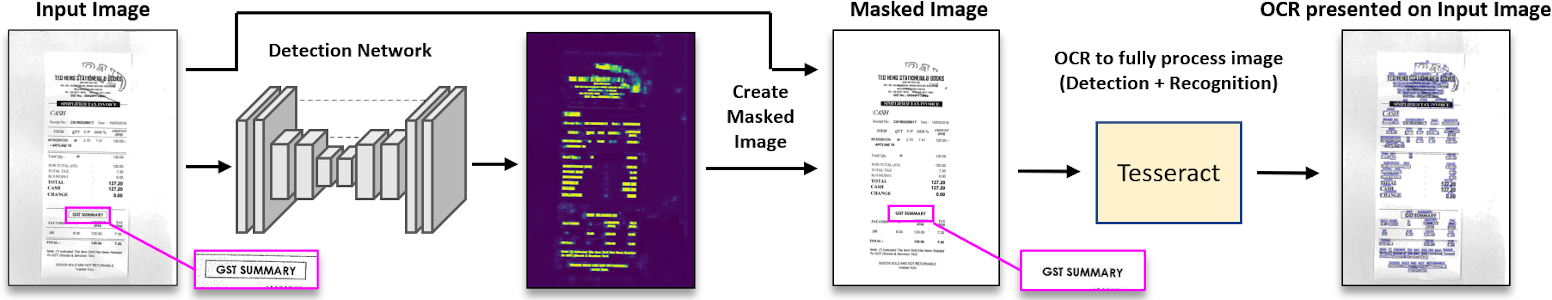}
  \caption{A diagram of our system. The image is analyzed by the U-Net text detector and then undergoes a masking operation to eliminate non-textual artifacts. The clean document fed into the OCR technology results in superior text detection and recognition.}
  \label{fig:pipeline}
\end{figure*}

In this work we propose a system to improve OCR results on degraded documents.
Specifically, we create a pipeline which can be easily and readily applied to improve a standard OCR platform such as Tesseract.
The core concept is to apply a pre-processing step with a designated detection network to perform a masking operation before processing the document with Tesseract (see Figure \ref{fig:pipeline}).

It is important to note that the main purpose of using this pipeline is to utilize the OCR's ability to perform better recognition when working at the document level.
When using the same detector and applying OCR on every detection separately, the results are not as satisfactory due to the fact that contextual data could not be leveraged.
Indeed, even if the detection network within Tesseract's pipeline could be replaced with the detector used for masking, results would likely not be better, as leveraging the contextual data was learned from clean documents.
When the input is a noisy document, the artifacts given in the image will hamper leveraging the contextual element even when given a noise-free detector due to the appearance of the text's local visual surroundings.

To train a document text detector, we present a deep learning U-Net architecture \cite{unet} trained on our presented dataset.
The dataset is synthesized with a variety of noise and difficult backgrounds as well as novel hard-negative samples to promote training of robust text detectors.
We make the dataset publicly available as a standalone pre-generated archive of 100k documents.

We perform a series of evaluations demonstrating the usefulness of the dataset, our network, and the masking approach.
Of note is the fact we demonstrate the performance on SROIE \cite{sroie}, a publicly available dataset of scanned receipts which accurately represents documents under difficult conditions.

Our contributions are as follows:
\begin{itemize}
    \item We present our new masking formulation for improving OCR results on noisy documents.
    \item We present a U-Net architecture and training methodology, and demonstrate its usefulness for text detection in documents.
    \item We propose a new data synthesis approach with a novel hard-negative component, and make it publicly available \footnote{\url{https://github.com/ophirazulai/SyntheticNoisyDocsDataset}} as a pre-generated dataset for the problem of text detection, and show the benefit of training given this dataset.
\end{itemize}

\section{Previous Work}

\subsection{OCR}

For the task of OCR, the most prominent and wide-spread solution is Tesseract \cite{tesseract}.
Originally, Tesseract was released as open-source in 2005 using mainly classic computer vision techniques including edge connected components analysis, blob filtering, quadratic spline fit, and recognition using topological features, and polygonal approximation.
Starting from version 4, Tesseract employs LSTMs \cite{lstm1, lstm2} and the CTC loss \cite{ctc} which are the current state-of-the-art approaches for text recognition.

Tesseract is considered the state of the art for OCR with regard to a commodity which is broadly applicable and easy to operate.
The fact that it is so broadly used makes creating techniques to improve Tesseract's results doubly useful.
For this reason we present our masking technique as a general pre-processing step which can easily improve Tesseract results and can be readily utilized.

However, when time or resource constraints are not an issue, advanced methods for extracting text from images exist.
Natural Scene Text recognition (NST) networks have risen recently in popularity \cite{nst}, but these approaches are outside of the scope of this work which focuses on light-weight text extraction from documents.

In this work we focus mainly on using the proposed masking approach before applying Tesseract as an OCR engine.
However, the same reasoning and methodology can apply to other OCR systems.

\subsection{Detection}

For detecting multiple objects from images, a very common approach is creating segmentation maps using a U-Net architecture \cite{unet}.
The architecture is characterized by the convolution layers which condense the spatial element to a bottleneck, and then up-convolutions which return the semantic information to the original spatial dimensions.
There is a variety of uses for the U-Net architecture and variations, including image segmentation (with a wide usage in medical imaging) \cite{unet_med1, unet_med2, unet_med3}, but also in other tasks such as saliency detection \cite{unet_salient}, or as GAN descriminators \cite{unet_gan}. 

Many powerful text detectors are constructed with architectures to promote NST.
EAST \cite{east} features a contracting and expanding network similar to the U-Net, and performs regression on the quadrilaterals based on the feature which is also used to generate the score map.
PAN \cite{pan} also adopts a contracting and expanding network, and adds explicit kernel learning to isolate and better separate close text in the pixel aggregation stage.

\subsection{Masking}

In this work we use masking of non-textual elements as a type of de-noising technique to enable Tesseract to utilize contextual information without considering noise and artifacts \cite{context}.

The term masking can sometimes refer to attention modules \cite{attention1, attention2}, or as part of training transformer networks \cite{transformer}.
Since the methodology we adopt is to perform the masking as a pre-processing stage and to leave Tesseract as a black-box, the methods above are not equivalent to our masking action which does not integrate knowledge or share semantic information.

Straightforward de-noising methods try to model the noise and convert the document accordingly \cite{denoise1, denoise2, denoise3, noise}.
These methods do not rely on the use of a text detector which can be seen as an advantage, but often are limited to specific types of noise and can degrade the quality, sharpness, and shape of the actual textual elements.
In our approach, instead of trying to model the noise, we aim instead on learning the ability to isolate and detect the text regions despite the noise present.

Finally, a common way to improve OCR results without intervening in the detection and recognition process is through post-OCR error correction \cite{post1, post2}.
These steps often leverage language models and information, and also represent one of the types of context which Tesseract uses to improve OCR quality when performing on the document-level.
Typically the types of linguistic and context errors here do not overlap with the ones our masking approach tries to solve such as noise and non-text artifacts, therefore this domain is beyond the scope of our work.

\subsection{Document Datasets}

The availability of datasets to train document OCR is limited.

FUNSD \cite{funsd} consists of 199 documents with roughly 31k word annotations.
The tasks and goals presented with the FUNSD dataset include mainly spatial layout analysis and form understanding.
With quantity of this magnitude, this dataset can be useful for training and evaluating the tasks which require the component of semantic understanding.
However, for the task of text detection it is necessary to have a much larger collection to encompass the low-level variability which exists when attempting to isolate text shapes from non-ideal backgrounds.

SROIE \cite{sroie} is a dataset consisting of scanned receipts for the tasks of OCR and key information extraction.
Despite the large number of samples, the word count per document on these receipts is not large enough to promote training text detectors from scratch.
We do however leverage this dataset for evaluation in Section \ref{sec:evaluation}.

Some additional datasets are Brno \cite{dataset_brno} which includes mostly spatial and lighting variations, quality assessment \cite{dataset_quality1, dataset_quality2} which concentrate on motion and focus blur, SmartDoc \cite{dataset_smartdoc} which is comprised of videos and only 10 documents, and some others \cite{dataset1, dataset2, dataset3, dataset4}.
However, none of these contain enough data to reliably train a robust text detector for documents.

The exception to the above is DDI-100 \cite{ddi}.
This dataset includes 7000 documents which then undergo a variety of transformations.
Despite the strength of utilizing real documents, we evaluate and show in this work that the variability and distortions in the dataset are not diverse enough to train a powerful text detector which is truly robust to noise.

In contrast, NST datasets have risen greatly in popularity in the past years \cite{dataset_nst1, dataset_nst2, dataset_nst3, dataset_nst4, dataset_nst5, dataset_nst6, dataset_nst7, dataset_nst8}.
However, the challenges presented in these datasets including irregular fonts and artistic text shapes and layouts, do not correctly represent the types of situations that a document text detector needs to learn to overcome.

\section{Method}

\subsection{Dataset}

We now present our synthetic automatically generated dataset for text detection in documents.

We use a python framework with the PIL library to synthesize text on images.
Backgrounds are selected at various set probabilities with the options of white, natural image, and texture.
For the latter, the textures are converted to grayscale and then a contour filter and a random dynamic range pixel value stretch is applied.

Text is synthesized with font size ranging from 9 to 100 pixels and font randomly selected 80\% from 20-30 common fonts and 20\% from a large assortment of unique fonts.
The text to synthesize is selected randomly from a wikipedia content database \cite{wikipedia}, which includes a large corpus of words, numbers, domains, dates, phone numbers, URLs, and more. The ground truth heat-map is generated as character-level bounding boxes to avoid protruding letters causing the background around smaller letters to be labeled as text (see Figure \ref{fig:gt}).

Font-level noise is randomly added chosen from speckled dots, binarization, and random spatial distortions.
Random small rotations are added to represent miss-alignment for scanned documents.
As a final step, document-level noise is added randomly in the form of blur, compression, or downsampling.
This step represents expected distortions which are likely to appear during scanning or photographing a document.

\begin{figure}[t]
  \centering
  \includegraphics[width=0.45\linewidth]{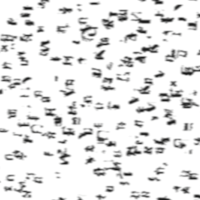}
  \hspace{1em}
  \includegraphics[width=0.45\linewidth]{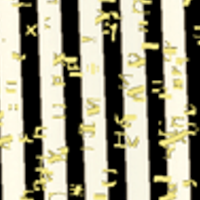}
  \caption{Examples of rendered hard-negatives. The detector learns to disregard character-like shapes and contours if they do not represent actual letters.}
  \label{fig:hard_negatives}
\end{figure}

To promote powerful negative sample filtering, we present a novel hard-negative synthesis approach to create particularly difficult data with which to train the detection network (see Figure \ref{fig:hard_negatives}).
Characters are generated and cropped into quarter-sized segments.
A crop is augmented by a random rotation and scaling and re-rendered to represent contours which are character-like but not actual distinguishable letters.

\subsection{U-Net Network}

We adopt a classic U-Net architecture for our text detection network.
Four layers of convolution and then up-convolution are performed with 32, 64, 128, and 256 channels.
In the up-convolution process skip-connections are employed by concatenating the output feature  of the up-convolution with the feature of the regular convolution at the same level.

\begin{figure}[b]
  \centering
  \includegraphics[width=\linewidth]{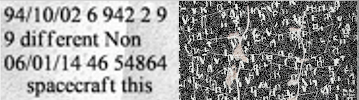}
  \includegraphics[width=\linewidth]{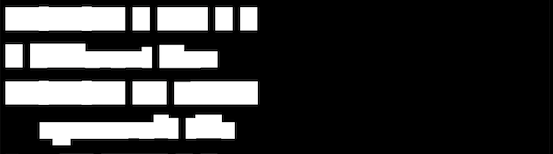}
  \caption{A miniature example of a generated document with only two synthesized tiles. Below is the matching generated binary ground-truth map for text localization.}
  \label{fig:gt}
\end{figure}

We use 100k synthesized documents 1024x1024 pixels, each consisting of a random number of synthesized tiles in each document (see Figure \ref{fig:gt}).
The dice loss \cite{dice} is used with the ADAM optimizer and a learning rate of 1e-5.

\begin{table*}[t]
\begin{center}
\caption{Visual results on two images from the SROIE dataset.}
\label{tab:visual_results}
\begin{tabular}{ccc}
& 						Original&	Masked\\
&			\includegraphics[width=0.15\linewidth]{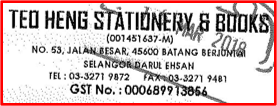}&			\includegraphics[width=0.15\linewidth]{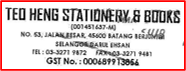}\\
\rotatebox[origin=c]{90}{X51005361908}&			\raisebox{-.5\height}{\includegraphics[width=0.15\linewidth]{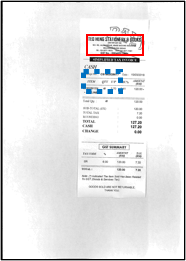}}&			\raisebox{-.5\height}{\includegraphics[width=0.15\linewidth]{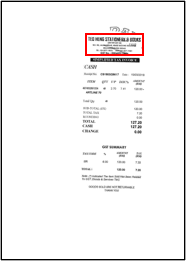}}\\
\rotatebox[origin=c]{90}{X00016469669}&			\raisebox{-.5\height}{\includegraphics[width=0.15\linewidth]{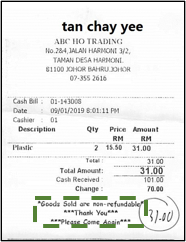}}&			\raisebox{-.5\height}{\includegraphics[width=0.15\linewidth]{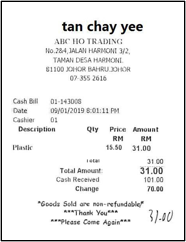}}
\end{tabular}
\begin{tabular}{ccc}
& 						Without Masking&	With Masking\\
\cite{east}&		\raisebox{-.5\height}{\includegraphics[width=0.25\linewidth]{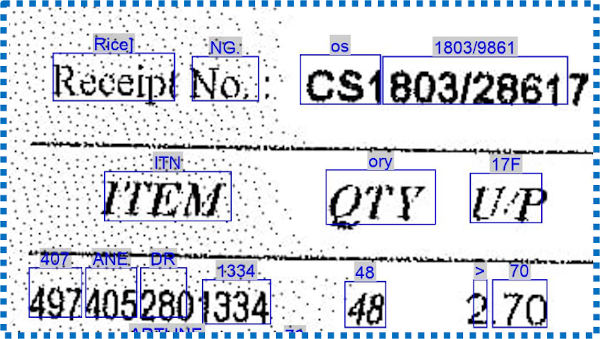}}&		\raisebox{-.5\height}{\includegraphics[width=0.25\linewidth]{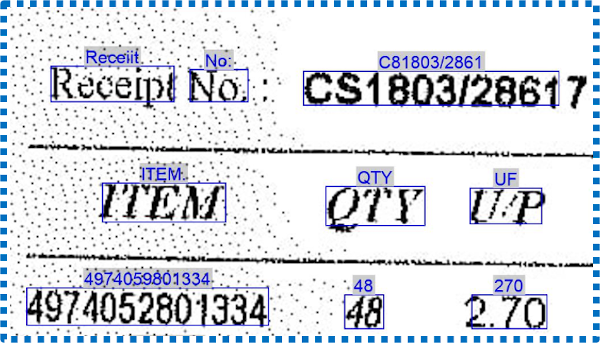}}\\
Ours&			\raisebox{-.5\height}{\includegraphics[width=0.25\linewidth]{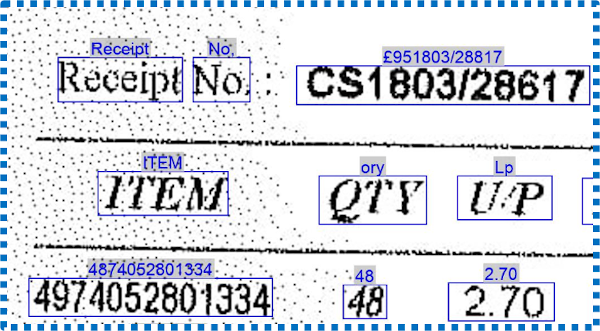}}&		\raisebox{-.5\height}{\includegraphics[width=0.25\linewidth]{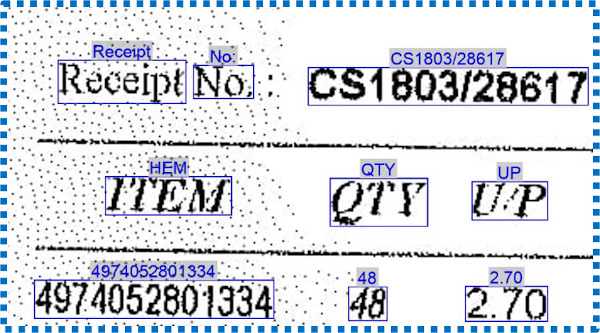}}\\
\cite{east}&		\raisebox{-.5\height}{\includegraphics[width=0.25\linewidth]{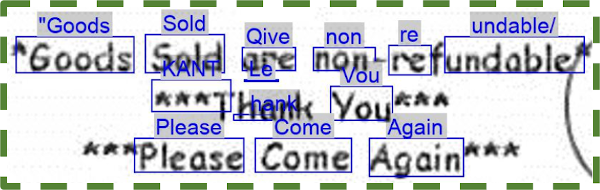}}&		\raisebox{-.5\height}{\includegraphics[width=0.25\linewidth]{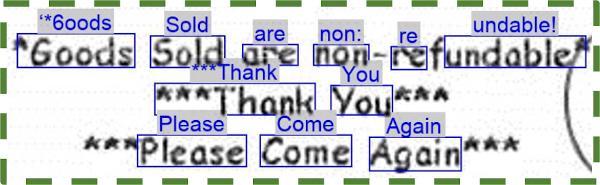}}\\
Ours&			\raisebox{-.5\height}{\includegraphics[width=0.25\linewidth]{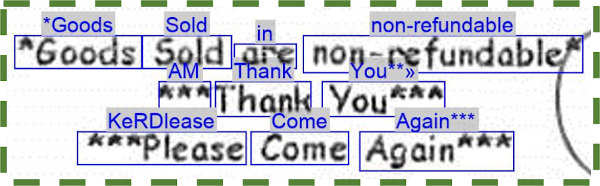}}&		\raisebox{-.5\height}{\includegraphics[width=0.25\linewidth]{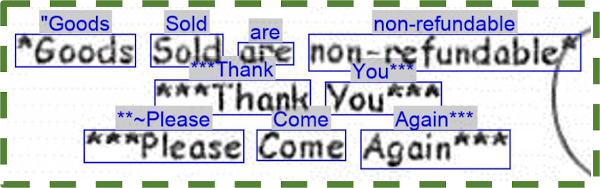}}
\end{tabular}
\end{center}
\end{table*}

\subsection{Detection Masking}

The output of the text detector is used to mask the given image.
Areas which were not detected as containing text are blanked out and then the cleaned image is fed into Tesseract.
Tesseract is run on the document-level, without providing the detections from the external text detector.

We note that a specific weakness of Tesseract is to sometimes identify entire paragraphs or lines as a single detection which impedes recognition.
Given the external text detection results, a post-processing procedure can be performed where these large detections are identified, merged with the text detector, and then resent to Tesseract for re-processing as individual detections.

\begin{table}[b]
\begin{center}
\caption{Results on SROIE dataset. Detection measured by F1-score, and recognition measured by average case-insensitive Edit Score (ES). `Recognition' indicates using the method's detections for word-level recognition. `Masked' indicates using our masking method and document-level recognition.}
\label{tab:results}
\begin{tabular}{c||c|cc}
& 						Detection&	Recognition&		Recognition\\
Method&					F1&			ES&			ES Masked\\
\hline
Tesseract&				84&			\multicolumn{2}{c}{67}\\
PAN \cite{pan}&			75&			37&					51\\
EAST \cite{east}&		84&			48&					59\\
U-Net - DDI \cite{ddi}&	82&			56&					62\\
U-Net - Ours&			92&			72&					75
\end{tabular}
\end{center}
\end{table}

\section{Evaluation}
\label{sec:evaluation}

We evaluate our masking technique, detection network, and contribution of our dataset on the SROIE dataset \cite{sroie}.
We used all 1064 images, with annotated text bounding boxes and transcribed words.

In Table \ref{tab:results} we present the results of our evaluation.
We measure detection using the F1-score of correctly detected bounding boxes using IOU 0.5.
We measure recognition by the normalized edit distance for an entire document, where for True-Positives we calculate the edit distance as the Levenshtein distance, and for False-Positives and False-Negatives we set the edit distance as the length of the string.
The normalized edit distance is the sum of edit-distances divided by the length of the text, and the Edit Score (ES) is 1 minus the edit distance.
Case is rendered insensitive as the annotations of the SROIE dataset do not include upper or lower case.

Tesseract results act as the baseline where the OCR is run at the document level.
The consistent improvement which can be seen for all methods when using the masking technique shows the generality and usefulness of the approach.
The improvement for U-Net between `DDI' and `Ours' shows the benefit of using out training dataset for the  task of text detection.

In Table \ref{tab:visual_results} we show some visual results from the SROIE dataset.
On the left we show visually the output of the masking operation which results in a cleaner and often more eligible document.
On the right we show close-up examples on the original document with the detection and recognition results visually embedded.
`Without Masking' represents using the detections and applying Tesseract on the word-level, while `With Masking' represents using the masking technique and applying Tesseract on the cleaned output at the document-level.

\section{Conclusions}

In this work we presented a masking technique based on a designated text detector to improve document OCR.
We introduced our new synthesized dataset with a novel hard-negative component designed to empower robust detection.
Finally, through evaluation we showed the benefits of using the masking approach and of using the dataset to utilize OCR performance which utilizes contextual data on documents.

\bibliographystyle{ACM-Reference-Format}
\bibliography{my_bib}

\end{document}